\def\year{2018}\relax
\documentclass[letterpaper]{article} 
\usepackage{aaai18}  
\usepackage{times}  
\usepackage{helvet}  
\usepackage{courier}  
\usepackage{url}  
\usepackage{graphicx}  
\usepackage{caption}

\usepackage{natbib}
\usepackage{algorithm}
\usepackage{algorithmic}

\usepackage{amsmath}
\usepackage{amsthm}
\usepackage{amssymb}

\begin{document}
%
\title{Improving Bi-directional Generation \\ between Different Modalities with Variational Autoencoders}
\author{Masahiro Suzuki\thanks{masa@weblab.t.u-tokyo.ac.jp}, Kotaro Nakayama, Yutaka Matsuo\\
The University of Tokyo, Japan
}
\maketitle
\begin{abstract}
We investigate deep generative models that can exchange multiple modalities bi-directionally, e.g., generating images from corresponding texts and vice versa. 
A major approach to achieve this objective is to train a model that integrates all the information of different modalities into a joint representation and then to generate one modality from the corresponding other modality via this joint representation. We simply applied this approach to variational autoencoders (VAEs), which we call a joint multimodal variational autoencoder (JMVAE).
However, we found that when this model attempts to generate a large dimensional modality missing at the input, the joint representation collapses and this modality cannot be generated successfully. Furthermore, we confirmed that this difficulty cannot be resolved even using a known solution.
Therefore, in this study, we propose two models to prevent this difficulty: JMVAE-kl and JMVAE-h. 
Results of our experiments demonstrate that these methods can prevent the difficulty above and that they generate modalities bi-directionally with equal or higher likelihood than conventional VAE methods, which generate in only one direction. Moreover, we confirm that these methods can obtain the joint representation appropriately, so that they can generate various variations of modality by moving over the joint representation or changing the value of another modality. 
\end{abstract}

\section{Introduction}
\label{Introduction}

Information in our world is represented through various modalities. Although images are represented by pixel information, these can also be described with text or tag information. People often exchange such information bi-directionally. For instance, we can not only imagine what a ``young female with a smile who does not wear glasses'' looks like, but we can also add this caption to a corresponding photograph. Our objective is to design a model that can exchange different modalities bi-directionally like people. We call this ability bi-directional generation.

\begin{figure}[tbh]
 \begin{center}
  \includegraphics[scale=0.9]{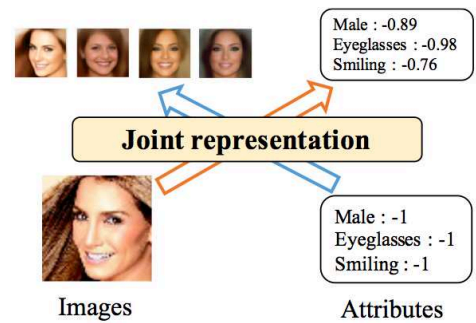}
 \end{center}
 \caption{Example of bi-directional generation between different modalities via a joint representation with JMVAE-kl, one of our proposed models.}
 \label{fig:overview} 
\end{figure}

Each modality typically has a different kind of dimension and structure, e.g., images (real-valued and dense) and text (discrete and sparse). Therefore, the relations among modalities might have high nonlinearity. To discover such relations, deep neural network architectures have been used widely \citep{Ngiam2011a, Srivastava2012}.

A major approach to realizing bi-directional generation between modalities is to train the network architecture that shares the top of hidden layers in modality specific networks \citep{Ngiam2011a}. The salient advantages of this approach are that the model with multiple modalities can be trained end-to-end, and that the trained model can extract a joint representation, which is a more compact representation integrating all modalities. The model can easily generate one modality from another modality via this representation if it can obtain the joint representation properly. 
Another simple approach might be to create networks for each direction and to train them independently. Several models have been proposed to generate modalities in one direction \citep{Kingma2014,Sohn2015,Pandey2016}. However, when generating modalities bi-directionally, the number of required networks is expected to increase exponentially as the modality increases. Moreover, the hidden layers of each network would not be synchronized during training. Therefore, this simple approach is not efficient to realize our objective.

To generate different modalities, it is important to model the joint representation as probabilistic latent variables. This is because, as described above, different modalities have different structures and dimensions, so their relation should not be deterministic. The best known approach by this probabilistic manner is to use deep Boltzmann machines (DBMs) \citep{Srivastava2012, Sohn2014}. However, it is computationally difficult for DBMs to train especially high-dimensional data because of MCMC training. 

Variational autoencoders (VAEs) \citep{Welling2014,Rezende2014}, a deep generation model, have an advantage that they can handle higher-dimensional datasets than DBMs because back-propagation can be used to train them. Therefore, we extend VAEs to a model that be able to generate modalities bi-directionally. This extension method is extremely simple: as with previous neural networks and DBMs approaches, latent variables of generative models corresponding to each modality are shared. We call this model a joint multimodal variational autoencoder (JMVAE). However, results show that if we miss the input of the high-dimensional modality we want to generate, the latent variable, i.e. the joint representation, of JMVAE collapses and this modality cannot be generated successfully. Although a method for addressing this difficulty has been proposed \citep{Rezende2014}, we demonstrate that this method cannot resolve this difficulty when the missing modality has higher dimensions than another one.

Therefore, we propose two new models to address difficulty presented above: JMVAE-kl and JMVAE-h. JMVAE-kl takes an approach of preparing new encoders with one input for each modality apart from the encoder of JMVAE and reducing the divergence between them. By contrast, JMVAE-h makes its latent variable a stochastic hierarchical structure to prevent its collapse. Figure \ref{fig:overview} shows that JMVAE-kl can generate modalities bi-directionally with different dimensions such as images and attributes. 

The main contributions of this paper are described below.
\begin{itemize}
\item We present a simple extension of VAEs to generate modality bi-directionally, which we call JMVAE. However, JMVAE cannot generate a high-dimensional modality well if the input of this modality is missing, and a known method of solution cannot resolve this issue.
\item We propose two models, JMVAE-kl and JMVAE-h, which prevent a latent variable from collapse when a high-dimensional modality is missing. We confirm experimentally that this method resolves this issue.
\item We demonstrate that these methods can generate modalities similarly or more properly than conventional VAEs that generate in only one direction. 
\item We demonstrate that they can appropriately obtain the joint representation containing different modality information, which shows that they can generate various variations of modality by moving over these latent variables or changing the value of another modality.
\end{itemize}

\section{Related work} 

A common approach to dealing with multiple modalities in deep neural networks is to share the top of hidden layers in modality-specific networks. \citet{Ngiam2011a} proposes this approach with deep autoencoders, which revealed that it can extract better representations than single-modality settings can. Actually, \citet{Srivastava2012} adapts this idea to work with deep Boltzmann machines (DBMs) \citep{salakhutdinov2009deep}, which are generative models with undirected connections based on maximum joint likelihood learning of all modalities. Therefore, this model can generate modalities bi-directionally. Another study \citet{Sohn2014} improves this model to exchange multiple modalities effectively, which are based on minimizing the variation of information. 

Recently, VAEs \citep{Welling2014,Rezende2014} have been used to train such high-dimensional modalities. Several studies \citep{Kingma2014,Sohn2015} have examined the use of conditional VAEs (CVAEs), which maximize a conditional log-likelihood by variational methods. In fact, many studies are based on efforts with CVAEs to train various multiple modalities such as handwriting digits and labels \citep{Kingma2014,Sohn2015}, object images and degrees of rotation \citep{kulkarni2015deep}, facial images and attributes \citep{Larsen2015, yan2015attribute2image}, and natural images and captions \citep{mansimov2015generating}. The main features of CVAEs are that the relation between modalities is one-way and a latent variable does not include information of a conditioned modality\footnote{According to \citep{Louizos2015}, this independence might not be satisfied strictly because the encoder in CVAEs still has dependence.}, which is an unsuitable characteristic for our objective. \citet{Pandey2016} proposes the use of a conditional multimodal autoencoder (CMMA), which also maximizes the conditional log-likelihood but which makes this latent variable connected directly from a conditional variable, i.e., these variables are not independent. However, CMMA still considers that these modalities are generated in a single fixed direction. 

Another well-known approach of deep generation models is generative adversarial nets (GANs) \citep{Goodfellow2014}. Even in the case of GANs, when handling multiple modalities, it is often modeled by generation in one direction such as conditional GANs \citep{mirza2014conditional}.

Recently, GANs that can generate images bi-directionally from images have been proposed \citep{liu2016coupled, liu2017unsupervised, zhu2017unpaired}. These models train perfect pixel-by-pixel correspondence between modalities of the same dimension and intentionally ignore probabilistic factors. However, no complete correspondence exists between modalities of different kinds that we examine specifically in this study. Our methods can train the probabilistic joint representation integrating all modality information, so they can obtain a probabilistic relation between modalities.

\begin{figure*}[tb]
 		\begin{center}
		\includegraphics[scale=0.8]{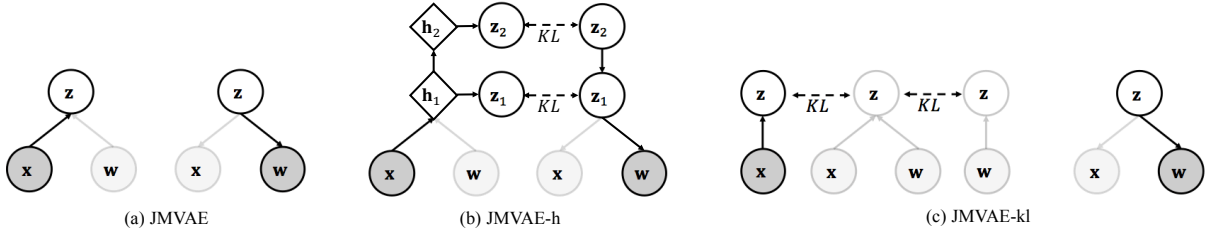}	
		\end{center}	
			\caption{Inference (or encoder, left figures) and generative (or decoder, right figures) distributions for (a) JMVAE (b) JMVAE-h, and (c) JMVAE-kl. These figures reflect how $q(\mathbf{z}|\mathbf{x})$ and $p(\mathbf{w}|\mathbf{z})$ are modeled on each approach. Circles represent stochastic variables. Diamonds represent deterministic variables.}
		\label{fig:approaches}
\end{figure*}

\section{VAEs with multiple modalities}
This section first briefly presents the formulation of VAEs. Subsequently, a simple extension of VAEs to multiple modalities is introduced.

\subsection{Variational autoencoders}
Given observation variables $\mathbf{x}$ and corresponding latent variables $\mathbf{z}$, their generating processes are definable as $\mathbf{z}\sim p(\mathbf{z})=\mathcal{N}(\mathbf{0},\mathbf{I})$ and $\mathbf{x}\sim p_\theta(\mathbf{x}|\mathbf{z})$, where $\theta$ is a model parameter of $p$. The objective of VAEs is maximization of the marginal distribution $p(\mathbf{x})=\int p_\theta(\mathbf{x}|\mathbf{z})p(\mathbf{z})d\mathbf{x}$. Because this distribution is intractable, we instead train the model to maximize the following lower bound.
\begin{eqnarray}
\log p(\mathbf{x}) &\geq& -D_{KL}(q_\phi(\mathbf{z}|\mathbf{x})||p(\mathbf{z})) + E_{q_\phi(\mathbf{z}|\mathbf{x})}[\log p_\theta(\mathbf{x}|\mathbf{z})] \nonumber \\ 
&\equiv& \mathcal{L}_{VAE}(\mathbf{x}), 
\label{eq:VAE_lower}
\end{eqnarray}
where $q_\phi(\mathbf{z}|\mathbf{x})$ is an approximate distribution of posterior and $\phi$ is a model parameter of $q$. We designate $q_\phi(\mathbf{z}|\mathbf{x})$ as {\rm encoder} and $p_\theta(\mathbf{x}|\mathbf{z})$ as {\rm decoder}. Moreover, in Eq. \ref{eq:VAE_lower}, the first term represents a regularization. The second one represents a negative reconstruction error.

To optimize the lower bound $\mathcal{L}(\mathbf{x})$ with respect to parameters $\theta,\phi$, we estimate the gradients of Eq. \ref{eq:VAE_lower} using stochastic gradient variational Bayes (SGVB). If we consider $q_\phi(\mathbf{z}|\mathbf{x})$ as a Gaussian distribution $\mathcal{N}(\mathbf{z};{\boldsymbol \mu},{\rm diag} ({\boldsymbol \sigma}^2))$, where $\phi=\{{\boldsymbol \mu},{\boldsymbol \sigma}^2\}$, then we can reparameterize $\mathbf{z}\sim q_\phi(\mathbf{z}|\mathbf{x})$ to $\mathbf{z}={\boldsymbol \mu}+{\boldsymbol \sigma} \odot{\boldsymbol \epsilon}$, where ${\boldsymbol \epsilon}\sim \mathcal{N}(\mathbf{0},\mathbf{I})$. Therefore, we can estimate the gradients of the negative reconstruction term in Eq. \ref{eq:VAE_lower} with respect to $\theta$ and $\phi$ as $\nabla_{\theta, \phi} E_{q_\phi(\mathbf{z}|\mathbf{x})}[\log p_\theta(\mathbf{x}|\mathbf{z})] = E_{\mathcal{N}({\boldsymbol \epsilon};\mathbf{0},\mathbf{I})}[\nabla_{\theta, \phi}\log p_\theta(\mathbf{z}|{\boldsymbol \mu}+{\boldsymbol \sigma}\odot {\boldsymbol \epsilon})]$. The gradients of the regularization term are solvable analytically. Therefore, we can optimize Eq. \ref{eq:VAE_lower} using standard stochastic optimization methods.

\subsection{Joint Multimodal Variational Autoencoders}

Next, we examine the $i.i.d.$ dataset $(\mathbf{X}, \mathbf{W})=\{(\mathbf{x}_1,\mathbf{w}_1)...,(\mathbf{x}_N,\mathbf{w}_N)\}$, where two modalities $\mathbf{x}$ and $\mathbf{w}$ have dimensions and structures of different kinds\footnote{In our experiment, they depend on the dataset, see the section of setting.}. We assume that these generative models are conditionally independent of the same latent variable $\mathbf{z}$, i.e. joint representation. It means that the latent variables of generative models corresponding to each modality are shared. Therefore, their generative process becomes $\mathbf{z} \sim p(\mathbf{z})$ and $\mathbf{x},\mathbf{w} \sim p(\mathbf{x},\mathbf{w}|\mathbf{z})=p_{\theta_\mathbf{x}}(\mathbf{x}|\mathbf{z})p_{\theta_\mathbf{w}}(\mathbf{w}|\mathbf{z})$, where $\theta_\mathbf{x}$ and $\theta_\mathbf{w}$ respectively represent the model parameters of each independent $p$. 


Considering an approximate posterior distribution as $q_\phi(\mathbf{z}|\mathbf{x},\mathbf{w})$, we can estimate a lower bound of the log-likelihood $\log p(\mathbf{x},\mathbf{w})$, as shown below.
\begin{align}
& \mathcal{L}_{JM}(\mathbf{x},\mathbf{w}) \nonumber \\
& = -D_{KL}(q_\phi(\mathbf{z}|\mathbf{x},\mathbf{w})||p(\mathbf{z})) \nonumber \\ 
&\;\;\;\; + E_{q_\phi(\mathbf{z}|\mathbf{x},\mathbf{w})}[\log p_{\theta_\mathbf{x}}(\mathbf{x}|\mathbf{z})] + E_{q_\phi(\mathbf{z}|\mathbf{x},\mathbf{w})}[\log p_{\theta_\mathbf{w}}(\mathbf{w}|\mathbf{z})].
\label{eq:JMVAE_lower}
\end{align}

Eq. \ref{eq:JMVAE_lower} has two negative reconstruction terms that are correspondent to each modality. As with VAEs, we designate $q_\phi(\mathbf{z}|\mathbf{x},\mathbf{w})$ as the encoder and both $p_{\theta_\mathbf{x}}(\mathbf{x}|\mathbf{z})$ and $p_{\theta_\mathbf{w}}(\mathbf{w}|\mathbf{z})$ as decoders. This model is a simple extension of VAEs to multiple modalities, and we call it a joint multimodal variational autoencoder (JMVAE). Because each modality has different feature representation, we set different networks for each decoder. 

\section{Complement of a missing modality}
\label{sec:inf_rec}

After training JMVAE, we can extract a joint latent representation by sampling from the encoder $q_{\phi}(\mathbf{z}|\mathbf{x},\mathbf{w})$ at testing time. Our objective is to exchange modalities bi-directionally, e.g., images to text and vice versa. In this setting, modalities that we want to sample are expected to be missing, so that inputs of such modalities are set to zero or random noise (Figure \ref{fig:approaches}(a)). In discriminative multimodal settings, this is a common means of estimating modalities from other modalities \citep{Ngiam2011a}, but it is difficult to handle this missing modality properly.

In VAE, \citet{Rezende2014} proposes a sophisticated approach of iterative sampling by Markov chain with a transition kernel to complement the missing value of input. In the case of JMVAE, the transition kernel when $\mathbf{x}$ is missing is $T(\tilde{\mathbf{x}}|\mathbf{x},\mathbf{w}) = \int p(\tilde{\mathbf{x}}|\mathbf{z})q(\mathbf{z}|\mathbf{x},\mathbf{w}) d\mathbf{z}$.

We can now estimate a missing modality $\mathbf{x}$ by first setting the initial value of $\mathbf{x}$ to random noise such as $\mathbf{x}\sim p(\mathbf{x})$. Then we conduct iterative sampling following to the above kernel. As the number of iterations increases, we can estimate complemented values better. As described in this paper, we call it the iterative sampling method.

However, if missing modalities are high-dimensional and complicated such as natural images compared to other modalities, then the inferred latent variable might collapse and generated samples might become incomplete. In the experiment, we will show that this difficulty cannot be prevented even using the iterative sampling method.

To resolve this issue, we propose two new models: JMVAE-h and JMVAE-kl.

\subsection{JMVAE-h}
Recently, some studies have extended the latent variables of VAEs to a stochastic hierarchical structure to improve the expressiveness and likelihood of models \citep{Burda2015, Sonderby2016a, gulrajani2016pixelvae}. Such a hierarchical structure of latent variables becomes robust against a missing input, so it might contribute to preventing them from collapse. 

We can extend JMVAE to a hierarchical structure of latent variables. Let the latent variable be the stochastic hierarchy of the $L$ layers $\mathbf{z}_{1},...,\mathbf{z}_{L}$\footnote{Note that the stochastic layers differ from the deterministic layers of neural networks.}, the joint distribution of JMVAE becomes $p(\mathbf{x},\mathbf{w}) = \int ... \int p_\theta(\mathbf{z}_L)p_\theta(\mathbf{z}_{L-1}|\mathbf{z}_L) ... p_\theta(\mathbf{x}|\mathbf{z}_1)p_\theta(\mathbf{w}|\mathbf{z}_1) d\mathbf{z}_1 ... d\mathbf{z}_L$, where all conditional distributions $p_\theta(\mathbf{z}_{l-1}|\mathbf{z}_l)$ are Gaussian and are parameterized by deep neural networks.

Various means are available to decompose the approximate distribution $q(\mathbf{z}|\mathbf{x},\mathbf{w})$. We follow \citep{gulrajani2016pixelvae}, which is $q(\mathbf{z}_1,...,\mathbf{z}_L|\mathbf{x},\mathbf{w}) = q_{\phi}(\mathbf{z}_1|\mathbf{x},\mathbf{w})\dots q_{\phi}(\mathbf{z}_L|\mathbf{x},\mathbf{w})$, where all conditional distributions $q_\phi(\mathbf{z}_{l}|\mathbf{x},\mathbf{w})$ are Gaussian. In this study, as with \citep{gulrajani2016pixelvae}, the structure from the input to the final stochastic layer consists of deterministic mappings  (each mapping is parameterized in a deep neural network), and the probabilistic output $\mathbf{z}_l$ of each stochastic layer is obtained from a deterministic output $\mathbf{h}_l$ (see Figure \ref{fig:approaches}(b)).

Therefore, the lower bound of JMVAE with a stochastic hierarchical structure becomes
\begin{align}
&\mathcal{L}_{JM_{h}}(\mathbf{x},\mathbf{w}) \nonumber \\
&= -\sum_{l=1}^L E_{q_\phi(\mathbf{z}_{l+1}|\mathbf{x},\mathbf{w})}[D_{KL}(q_\phi(\mathbf{z}_l|\mathbf{x},\mathbf{w})||p_\theta(\mathbf{z}_l|\mathbf{z}_{l+1}))] \nonumber \\ 
&\;\;\;\; + E_{q_\phi(\mathbf{z}_1|\mathbf{x},\mathbf{w})}[\log p_{\theta}(\mathbf{x}|\mathbf{z}_1)] + E_{q_\phi(\mathbf{z}_1|\mathbf{x},\mathbf{w})}[\log p_{\theta}(\mathbf{w}|\mathbf{z}_1)]. \nonumber \\
\end{align}

We call this model JMVAE-h, and demonstrate that it can prevent the issue of missing modality through experiments. In our experiments, we set $L=2$.
Figure \ref{fig:approaches}(b) shows the flow of generating $\mathbf{w}$ from $\mathbf{x}$ with JMVAE-h.

\subsection{JMVAE-kl}
\label{sec:JMVAE-kl}
In the stochastic hierarchical approach described above, the iterative sampling method is still important for generating missing modalities. However, it takes time to generate high-dimensional samples. Therefore, we propose JMVAE-kl as a model to generate appropriate missing samples without using the iterative sampling method.

Assume that we have encoders with a single input, $q_{\lambda}(\mathbf{z}|\mathbf{x})$ and $q_{\lambda}(\mathbf{z}|\mathbf{w})$, where $\lambda$ is a parameter, then we would like to train them by reducing the divergence between their encoders and an encoder $q_{\phi}(\mathbf{z}|\mathbf{x},\mathbf{w})$ (see Figure \ref{fig:approaches}(c)). Therefore, the object function of JMVAE-kl becomes
\footnotesize
\begin{align}
& \mathcal{L}_{JM_{kl}}(\mathbf{x},\mathbf{w}) \nonumber \\
& = \mathcal{L}_{JM}(\mathbf{x},\mathbf{w}) \nonumber \\
&\;\;\;\; - [D_{KL}(q_\phi(\mathbf{z}|\mathbf{x},\mathbf{w})||q_{\lambda}(\mathbf{z}|\mathbf{x})) + D_{KL}(q_\phi(\mathbf{z}|\mathbf{x},\mathbf{w})||q_{\lambda}(\mathbf{z}|\mathbf{w}))].\;
\label{eq:new_lowerbound}
\end{align}
\normalsize

Actually, JMVAE-h and JMVAE-kl have both benefits and shortcomings. JMVAE-kl must prepare encoders for each modality apart from the original encoder, but JMVAE-h requires no preparation of any additional encoder. Conversely, although JMVAE-h must apply the iterative sampling method to generate a missing modality, JMVAE-kl does not need to do so. Therefore, JMVAE-h is effective when there are three or more modalities and the dimension of the missing modality is not so large; JMVAE-kl is effective when there are modalities of only two kinds and the missing modality is high-dimensional.

 \begin{figure*}[tb]
 \begin{center}
  \includegraphics[scale=0.9]{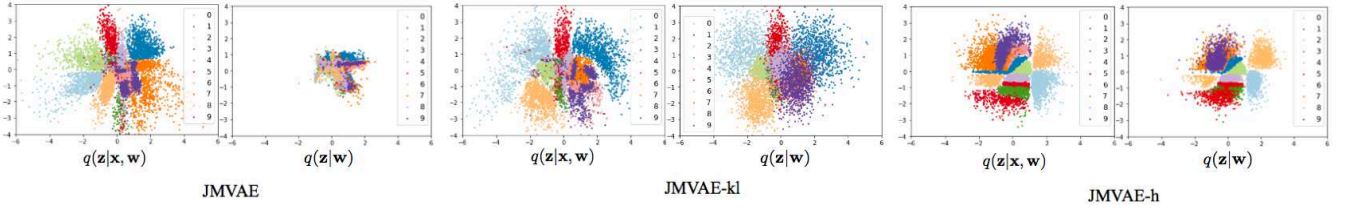}
 \end{center}
 \caption{Visualizations of 2-D latent representation. The number of iterative sampling on JMVAE and JMVAE-h is 10.}
 \label{fig:px_multimodal}
 \end{figure*}
 
 \begin{table}[tb]
  \begin{center}
  \small
  \begin{tabular}{lcc}
     Model & Label $\to$ Image & Image $\to$ Label \\\hline
     JMVAE & -977.2 & {\bf -0.2361} \\
     JMVAE-kl & {\bf -422.4} & -0.2628 \\
     JMVAE-h & -552.2 & -3.589 \\\hline
     CVAE & -448.8 & -5.293 \\
     CMMA & -451.1 & -0.2971 \\
  \end{tabular}
  
  \begin{tabular}{lcc}
     Model & Attributes $\to$ Image & Image $\to$ Attributes\\\hline
     JMVAE & -48763 & {\bf -43.97} \\
     JMVAE-kl & {\bf -6852} & -44.13 \\
     JMVAE-h & -7355 & -47.61 \\\hline
     CVAE & -6825 & -44.28 \\
     CMMA & -6920 & -44.57 \\
  \end{tabular}
  \end{center}
   \caption{Evaluation of the conditional log-likelihood. Models are trained and tested on MNIST (upper) and CelebA (lower).}   
  \label{tab:log-likelihood}   
\end{table}

\begin{figure}[tb]
 \begin{center}
  \includegraphics[scale=0.40]{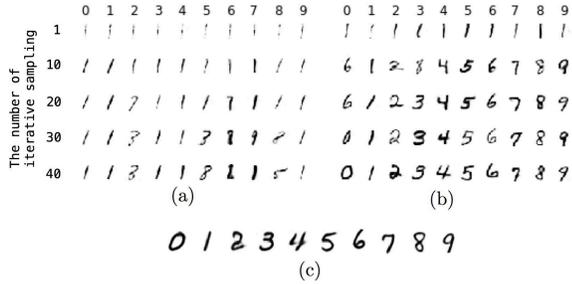}
 \end{center}
 \caption{Images ($\mathbf{x}$) generation from labels ($\mathbf{w}$) on the MNIST dataset. Each column corresponds to each element of space of $\mathbf{w}$, i.e., labels from 0 to 9. Going to the bottom of the row, the number of iterative sampling for generating $\mathbf{x}$ increases: (a) JMVAE, (b) JMVAE-h, and (c) JMVAE-kl.}
 \label{fig:generate_mnist}
 \end{figure}
 
 \begin{figure}[tb]
 \begin{center}
  \includegraphics[scale=0.40]{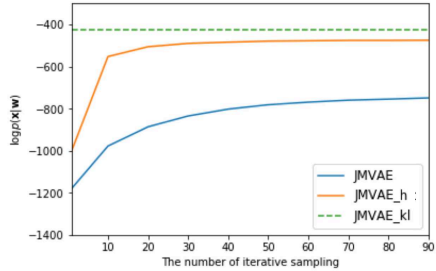}
 \end{center}
 \caption{Log-likelihood values for JMVAE, JMVAE-h, and JMVAE-kl model with different numbers of iterative sampling on the MNIST dataset.}
 \label{fig:plot_conditional_loglike}
 \end{figure}

\section{Experiments}
In this section, we confirm the following three points through experimentation: (1) The missing modality difficulty certainly occurs on JMVAE and our proposed methods can prevent this issue. (2) Our proposed models can generate modalities bi-directionally with the likelihood equivalent to (or higher than) models that generate modality in only one direction. (3) Our models can appropriately obtain the joint representation, which integrates information of different modalities.

\subsection{Datasets}
As described herein, we used two datasets: MNIST and CelebA \citep{Eyebrows2015}.

Originally, MNIST was not a dataset for multimodal setting. For this work, we used this dataset as a toy problem for various tests to verify of our model. We regard 784-dimensional handwriting images and corresponding 10 digit labels as two modalities. We used 50,000 as training set and the remaining 10,000 as a test set.

CelebA consists of 202,599 color facial images and 40 corresponding binary attributes such as male, eyeglasses, and mustache. For this work, we regard them as two modalities. This dataset is challenging because they have dimensions and structures of completely different kinds. Beforehand, we cropped the images to squares and resized them to 64 $\times$ 64 and normalized them. From the dataset, we chose 191,899 images that are identifiable facial images using OpenCV and used them for our experiment. We used 90\% out of all the dataset contents as a training set and the remaining 10\% of them as a test set.
\subsection{Settings}
\label{sec:model_arc}

For MNIST, we considered images as $\mathbf{x}\in \mathbb{R}^{784}$ and corresponding labels as $\mathbf{w}\in \{0,1\}^{10}$. We set $p(\mathbf{x}|\mathbf{z})$ as Bernoulli and $p(\mathbf{w}|\mathbf{z})$ as categorical distribution.
We used warm-up \citep{bowman2015generating, Sonderby2016a}, which first forces training only of the term of the negative reconstruction error and which then gradually increases the effect of the regularization term to prevent local minima during early training. We increased this term linearly during the first $N_t$ epochs as with \citet{Sonderby2016a}. Then we set $N_t=200$ and trained for $2000$ epochs on MNIST. Moreover, as described by \citet{Burda2015,Sonderby2016a}, we resampled the binarized training values randomly from MNIST images for each epoch to prevent over-fitting.

For CelebA, we considered facial images as $\mathbf{x}\in \mathbb{R}^{32 \times 32 \times 3}$ and corresponding attributes as $\mathbf{w}\in \{-1,1\}^{40}$.  
We set a Gaussian distribution for the decoder of both modalities, where the variance of the Gaussian was fixed to 1. We set $N_t=20$ and trained for $50$ epochs.

We used the Adam optimization algorithm \citep{Kingma2015a} with a learning rate of $10^{-3}$ on MNIST and $10^{-4}$ on CelebA. The models were implemented using Theano \citep{theTheanoDevelopmentTeam2016}, Lasagne \citep{Dieleman2015}, and Tars\footnote{https://github.com/masa-su/Tars}. See appendix for details of network architectures.

\subsection{Evaluation method}
For this experiment, we estimated the test conditional log-likelihood $\log p(\tilde{\mathbf{x}}|\mathbf{w})$ (or $\log p(\tilde{\mathbf{w}}|\mathbf{x})$) to evaluate the model performance. This estimate indicates how well a model can generate a modality from the corresponding another modality. Therefore, higher is better. We can estimate $\log p(\tilde{\mathbf{x}}|\mathbf{w})$ as
\footnotesize
\begin{align}
&\log p(\tilde{\mathbf{x}}|\mathbf{w}) \simeq \log \int q(\mathbf{z}|\mathbf{w})p(\tilde{\mathbf{x}}|\mathbf{z}) d\mathbf{z} \nonumber \\
&\;\;\;\;\;\;\;\;\;\;\;\;\;\;\;\;\; \geq \int q(\mathbf{z}|\mathbf{w})\log p(\tilde{\mathbf{x}}|\mathbf{z}) d\mathbf{z} \simeq  \frac{1}{N}\sum_{i=1}^N \log p(\tilde{\mathbf{x}}|\mathbf{z}^{(l)}),
\label{eq:conditional_log_likelihood}
\end{align}
\normalsize
where $\mathbf{z}^{(l)}\sim q(\mathbf{z}|\mathbf{w})$. In Eq. \ref{eq:conditional_log_likelihood}, we apply the sample approximation to the lower bound of the log-likelihood rather than to the log-likelihood directly. This is because the log-likelihood is biased\footnote{This lower bound is unbiased estimator. Moreover, \cite{Burda2015} shows that sample approximation of log-likelihood approaches the true log-likelihood as the number of samples increases.}. We set $N=10$ for all experiments.

To estimate the above, we should approximate $q(\mathbf{z}|\mathbf{w})$ (or $q(\mathbf{z}|\mathbf{x})$) and draw samples from them. The way of it depends on how to complement missing modalities. In the cases of JMVAE and JMVAE-h, we first set the input of modality which we want to generate as missing, and then apply iterate sampling in multiple times to complement the missing modality. This means that the estimation of log-likelihood on these models depends on the number of iterate sampling. Conversely, JMVAE-kl can estimate the approximate distribution $q(\mathbf{z}|\mathbf{w})$ (or $q(\mathbf{z}|\mathbf{x})$) directly at the training stage. For the approximation of conditional log-likelihood in JMVAE-h, see appendix.

\subsection{MNIST results}
\label{sec:quan}
 
\subsubsection{Confirmation of the missing modality difficulty}

Figure \ref{fig:generate_mnist}(a) presents results of generating $\mathbf{x}$ from $\mathbf{w}$. The top row, i.e. the image generated by application of iterative sampling only once, is blurred and is not properly generated. As the number of iterative sampling increases, the generated images become somewhat clearer,  but it is readily apparent that these images do not correspond to their labels. This result demonstrates that a missing modality cannot be generated even using the iterative sampling method. Figure \ref{fig:generate_mnist}(b) presents the result obtained in the case of JMVAE-h. Unlike the results presented in Figure \ref{fig:generate_mnist}(a), it is apparent that images corresponding to their labels are generated appropriately as the sampling number increases. Figure \ref{fig:generate_mnist}(c) is the result for JMVAE-kl. JMVAE-kl can generate digit images conditioned on numbers appropriately without iterating sampling.

Figure \ref{fig:plot_conditional_loglike} presents the conditional likelihood of each model under the number of iterative sampling changes. As the number of sampling increases, the log-likelihood increases for both JMVAE and JMVAE-h. However, in the case of normal JMVAE, its likelihood is much lower than that of JMVAE-kl no matter how much the number of sampling is increased. For JMVAE-h, the likelihood increases with a fewer sampling times than JMVAE, and the final likelihood becomes higher than normal JMVAE. On the other hand, in the case of JMVAE-kl, it can obtain high likelihood without iterative sampling.

\subsubsection{Evaluation of conditional log-likelihood}
This section describes evaluation of our models with conditional log-likelihood to evaluate the bi-directional generation of modalities quantitatively. Additionally, we compare the log-likelihood evaluation with conventional conditional VAEs, which are CVAE \citep{Kingma2014,Sohn2015} and CMMA \citep{Pandey2016}. Note that these models cannot generate modalities bi-directionally, so it is necessary to train them separately in each direction.

The upper of Table \ref{tab:log-likelihood} presents the test conditional log-likelihoods of JMVAE, our two improved models, and conventional conditional VAEs. The number of iterative sampling of JMVAE and JMVAE-h was set to 10. First, when generating an image from labels, JMVAE-kl and JMVAE-h improve their likelihood compared to JMVAE as expected. In addition, it turns out that these models have higher likelihood than existing VAEs modeling generation in only one direction. Next, in the case of generating labels from images, we find that there is not much difference in the likelihood for each model compared with generation in the opposite direction. This means that the missing modality difficulty does not occur when we generate a small dimensional modality from another large one. Note that the evaluation value of the conditional likelihood of JMVAE-h might be underestimated compared to the evaluation value of other models. This is because the approximation of conditional likelihood holds a lower bound than other models (see appendix for details). In light of this fact, we can see that both JMVAE-kl and JMVAE-h can be generated in both directions with the same or higher likelihood as that of conventional conditional VAEs.

\subsubsection{Visualization of the joint representation}

Figure \ref{fig:px_multimodal} presents a visualization of the joint representation in each model. For the case of JMVAE-h, we sampled at the top stochastic layer. Here, we sampled from the trained encoders with the dimensions of the latent variable set to 2.

Specifically examining the left of each model results in Figure \ref{fig:px_multimodal}, samples from all models in the latent space are distributed and labeled for each label, which indicates that the joint representation including two modalities is obtained. 

Next, we specifically examine the right in Figure \ref{fig:px_multimodal}. From JMVAE results, we can see that all the samples are distributed in a considerably small area irrespective of their labels. This result demonstrates that, if the image information is missing, the latent representation actually collapses. By contrast, the result of JMVAE-kl shows that it can obtain the latent representation that is almost unchanged from sampling from all modalities. Because samples are not gathered in a small area as JMVAE, JMVAE-kl prevents the latent representation from collapsing. Finally, regarding the result of JMVAE-h, they are distributed considerably separately for each label just like the result on the left figure. From this, it can be said that the collapse of latent representation is also prevented in JMVAE-h too.
 
 \begin{figure}[tb]
 \begin{center}
 \includegraphics[scale=0.60]{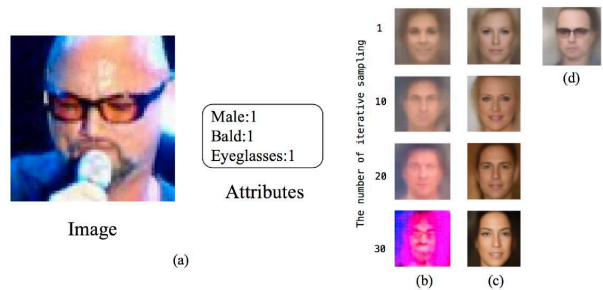}
 \end{center}
 \caption{Facial images ($\mathbf{x}$) generation from attributes ($\mathbf{w}$) on CelebA dataset. (a) is an example of the test set on CelebA. (b)-(d) are facial images which are generated from attributes of the example (a). Going to the bottom of the row, the number of iterative sampling for generating $\mathbf{x}$ increases: (b) JMVAE, (c) JMVAE-h, and (d) JMVAE-kl.}
 \label{fig:generate_celebA}
 \end{figure}
 
\begin{figure}[tb]
 \begin{center}
  \includegraphics[scale=0.43]{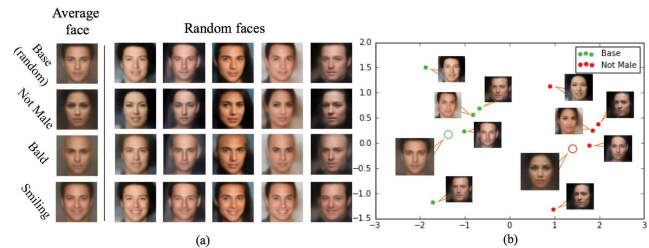}
 \end{center}
 \caption{(a) Generation of average faces and corresponding random faces. We first set all values of attributes randomly and designate them as Base. Then, we choose an attribute that we want to set (e.g., Male, Bald, Smiling) and change this value in Base to $2$ (or $-2$ if we want to set "Not"). Each column corresponds to the same attribute according to the legend. Average faces are generated from $p(\mathbf{x}|\mathbf{z}_{mean})$, where $\mathbf{z}_{mean}$ is a mean of $q(\mathbf{z}|\mathbf{w})$. Moreover, we can obtain various images conditioned on the same values of attributes such as $\mathbf{x}\sim p(\mathbf{x}|\mathbf{z})$, where $\mathbf{z}=\mathbf{z}_{mean}+{\boldsymbol \sigma}\odot{\boldsymbol \epsilon}$, ${\boldsymbol \epsilon}\sim \mathcal{N}(\mathbf{0},{\boldsymbol I})$. 
 Each row in random faces has the same ${\boldsymbol \epsilon}$. (b) PCA visualizations of latent representation. Colors show which attribute is conditioned on for each sample.} 
 \label{fig:mean_celeba}
\end{figure}
 
  \begin{figure*}[tb]
 \begin{center}
  \includegraphics[scale=0.6]{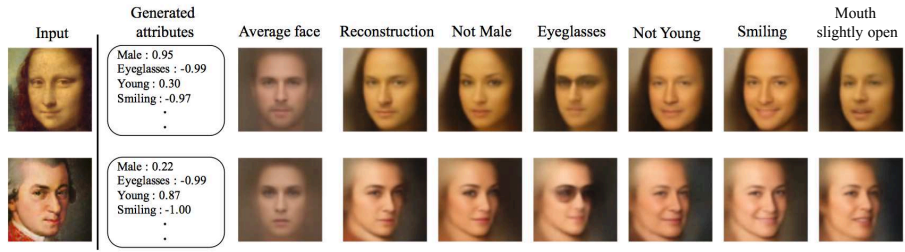}
 \end{center}
 \caption{Portraits of the Mona Lisa\protect\footnotemark (upper) and Mozart\protect\footnotemark  (lower), generated attributes and reconstructed images conditioned on varied attributes according to the legend. We cropped and resized it in the same way as CelebA. The procedure was the following: generate the corresponding attributes $\mathbf{w}$ from an unlabeled image $\mathbf{x}$. Generate an average face $\mathbf{x}_{mean}$ from the attributes $\mathbf{w}$. Select attributes that we want to vary and change the values of these attributes. Generate the changed average face $\mathbf{x}'_{mean}$ from the changed attributes. Obtain a changed reconstruction image $\mathbf{x}'$ by $\mathbf{x}+\mathbf{x}'_{mean}-\mathbf{x}_{mean}$.} 
 \label{fig:change_monalisa_celeba}
\end{figure*}
\footnotetext{https://en.wikipedia.org/wiki/Mona\_Lisa}
\footnotetext{https://en.wikipedia.org/wiki/Wolfgang\_Amadeus\_Mozart}
 
\subsection{CelebA results}

\subsubsection{Confirmation of the missing modality difficulty}
 
Figure \ref{fig:generate_celebA} presents the results obtained from generating facial images from attributes. From Figure \ref{fig:generate_celebA}(b), we can see that an ordinary JMVAE cannot generate an appropriate facial images at all from its attributes. In addition, unlike MNIST, it can be seen that the facial image largely collapses if the number of iterative sampling is increased. This suggests that the greater the difference in the amount of information between modalities becomes, the more severely the missing modality difficulty may become. On the other hand, it is confirmed that JMVAE-h improves the facial image quality, and that it can generate a more beautiful images when the number of iterative sampling is increased (Figure \ref{fig:generate_celebA}(c)). However, it can also be confirmed that these generated images do not correspond closely to the attributes. Specifically, no generated image shows a subject wearing eyeglasses, probably because the stochastic hierarchy deepened, the gap separating the modes corresponding to the attributes becomes smaller, which makes mixing in the latent space easier. Finally, for the case of JMVAE-kl, we can observe that facial images corresponding to attributes are generated by one sampling (Figure \ref{fig:generate_celebA}(d)).

\subsubsection{Evaluation of the conditional log-likelihood}
The lower of the Table \ref{tab:log-likelihood} shows the test conditional log-likelihoods of all models. The number of iterative sampling of JMVAE and JMVAE-h was set to 40. First, in the case of generating facial images from attributes, we can see that the likelihoods of JMVAE-kl and JMVAE-h are improved considerably compared to that of JMVAE. From this, we can see that even if the dimensions between the modalities are differ greatly, our improved models contribute to preventing the missing modality difficulty. Furthermore, as with the result of MNIST, we can see that our improved models can be generated in both directions with the same or higher likelihood as that of conventional VAEs.
 

\subsubsection{Generating faces from attributes and the joint representation on CelebA}

Next, we confirm that our model can obtain the joint representation appropriately.
Based on the results presented above, we used JMVAE-kl in the remaining experiments. Moreover, we combined JMVAE-kl with GANs to generate clearer images. We considered the network of $p(\mathbf{x}|\mathbf{z})$ as {\rm generator} in GANs. Then we optimized the GAN's loss with the lower bound of JMVAE-kl, which is the same approach to a VAE-GAN model \citep{Larsen2015}. See appendix for the network structure of the discriminator of this GAN.

Figure \ref{fig:mean_celeba}(a) portrays generated faces conditioned on various attributes. Results show that we can generate an average face for each attribute and various random faces conditioned on a certain attributes. Figure \ref{fig:mean_celeba}(b) shows that these samples are gathered for each attribute. In addition, in the group for each attribute, the average face is positioned almost at the center of each group, and random facial images are arranged around the average face. Furthermore, it can be confirmed that these arrangements are almost the same in both Base and Not Male. These results indicate that manifold learning of the joint representation works well.

\subsubsection{Bi-directional generation between faces and attributes on CelebA}
Finally, we demonstrate that our model can do bi-directionally generation between faces and attributes. Figure \ref{fig:change_monalisa_celeba} shows that JMVAE-kl can generate both attributes and changed images conditioned on various attributes from images that had no attribute information. These are possible because, as the result above shows, JMVAE-kl can properly obtain the joint representation integrating different modalities.

\section{Conclusion}
As described herein, we introduced the extension of VAEs to generate modalities bi-directionally, which we call JMVAE. However, results show that we cannot generate a high-dimensional modality properly if the input of this modality is missing, and that the known method, the iterative sampling method, cannot resolve this issue. We proposed two models, JMVAE-kl and JMVAE-h, to prevent the above missing modality difficulty. Results demonstrated that these proposed models prevent a latent variable from collapse and that they can generate modalities bi-directionally with equal or higher quality compared to models which can only generate them in one direction. Furthermore, because these methods appropriately obtain the joint representation, we found that samples of various variations can be generated as corresponding to a certain modality or a changed modality.


\bibliography{citation.bib}

\appendix

\section{The network architectures}

\subsection{Parameterization of distributions with deep neural networks}
The Gaussian distribution can be parameterized with deep neural networks as follows.
\begin{align}
&\mathcal{N}(\mathbf{z};{\boldsymbol \mu},{\rm diag} ({\boldsymbol \sigma}^2)),\nonumber \\
&{\boldsymbol \mu} = f_{\mu}(f_{\rm MLP}(\mathbf{x})), \nonumber \\
&{\boldsymbol \sigma}^2 = {\rm Softplus}(f_{{\sigma}^2}(f_{\rm MLP}(\mathbf{x}))),  \nonumber
\end{align} 
where $f_{\mu}$ and $f_{{\sigma}^2}$ are linear single layer neural networks and $f_{\rm MLP}$ means a deep neural network with arbitrary number of layers. Moreover, applying softplus function for each element of a vector is denoted as ${\rm Softplus}$.

The Bernoulli distribution is parameterized as
\begin{eqnarray}
p_\theta(\mathbf{x}|\mathbf{z})=\mathcal{B}(\mathbf{x};{\boldsymbol \mu}), {\boldsymbol \mu} = {\rm Sigmoid}(f_{{\mu}}(f_{\rm MLP}(\mathbf{z}))),\nonumber
\label{eq:Bern}
\end{eqnarray}
where ${\rm Sigmoid}$ is sigmoid function.

In the case of categorical distribution, we can parameterize it as
\begin{eqnarray}
p_\theta(\mathbf{x}|\mathbf{z})=\mathcal{C}(\mathbf{x};{\boldsymbol \mu}), {\boldsymbol \mu} = {\rm Softmax}(f_{{\mu}}(f_{\rm MLP}(\mathbf{z}))),\nonumber
\label{eq:cat}
\end{eqnarray}
where ${\rm Softmax}$ means softmax function.

\subsection{MNIST}

For the notation of model structures, we denote a linear fully-connected layer with $k$ units and ReLU as {\tt DkR}, and {\tt DkR} without ReLU as {\tt Dk}. In addition, the process of applying {\tt J} after {\tt I} is denoted as {\tt I-J}, and the process of concatenating the last layers of the two networks {\tt I, J} into one layer is denoted as {\tt (I,J)}.

Therefore, the network structures of encoders and decoders on MNIST are as follows.

\begin{itemize}
 \item $q(\mathbf{z}|\mathbf{x}, \mathbf{w})$ (Gaussian)
    \begin{itemize}
      \item $f_{\mu}$ and $f_{{\sigma}^2}$: {\tt D64}
      \item $f_{\rm MLP}$: {\tt (D512R-D512R, D512R-D512R)}
     \end{itemize}
 \item $q(\mathbf{z}|\mathbf{x})$, $q(\mathbf{z}|\mathbf{w})$, and $p(\mathbf{z}_1|\mathbf{z}_2)$ (Gaussian)
    \begin{itemize}
      \item $f_{\mu}$ and $f_{{\sigma}^2}$: {\tt D64}
      \item $f_{\rm MLP}$: {\tt D512R-D512R}
    \end{itemize}
 \item $p(\mathbf{x}|\mathbf{z})$ (Bernoulli)
    \begin{itemize}
      \item $f_{\mu}$: {\tt D784}
      \item $f_{\rm MLP}$: {\tt D512R-D512R}
     \end{itemize}   
 \item $p(\mathbf{w}|\mathbf{z})$ (Categorical)
    \begin{itemize}
      \item $f_{\mu}$: {\tt D10}
      \item $f_{\rm MLP}$: {\tt D512R-D512R}
     \end{itemize}             
 \item $q(\mathbf{z}_2|\mathbf{z}_1)$ (Gaussian)
    \begin{itemize}
      \item $f_{\mu}$ and $f_{{\sigma}^2}$: {\tt D64}
      \item $f_{\rm MLP}$: {\tt D512R-D512R-D64-D512R-D512R}
     \end{itemize}  
\end{itemize}

\subsection{CelebA}
We denote a convolutional layer (filter size : $4\times4$, number of channels : $k$, and stride : 2) with batch normalization and ReLU as {\tt CkBR}, and {\tt CkBR} without ReLU as {\tt CkB}. In addition, we denote a deconvolutional layer (filter size : $4\times4$, number of channels : $k$, and crop : 2) with batch normalization and ReLU as {\tt DCkBR}, {\tt DCkBR} without ReLU as {\tt DCkB}, and {\tt DCkB} without batch normalization as {\tt DCk}. Further, a linear fully-connected layer with $k$ units, ReLU, and batch normalization is denoted as {\tt DkBR} and a flatten layer is denoted as {\tt F}.

Using the above notations, the model structures on CelebA are as follows.

\begin{itemize}
 \item $q(\mathbf{z}|\mathbf{x}, \mathbf{w})$ (Gaussian)
    \begin{itemize}
      \item $f_{\mu}$ and $f_{{\sigma}^2}$: {\tt D128}
      \item $f_{\rm MLP}$: {\tt (C64R-C128BR-C256BR-C256BR-F, D512R-D512BR)-D1024R}
     \end{itemize}
 \item $q(\mathbf{z}|\mathbf{x})$ (Gaussian)
    \begin{itemize}
      \item $f_{\mu}$ and $f_{{\sigma}^2}$: {\tt D128}
      \item $f_{\rm MLP}$: {\tt C64R-C128BR-C256BR-C256BR-F-D1024R}
    \end{itemize}
 \item $q(\mathbf{z}|\mathbf{w})$ (Gaussian)
    \begin{itemize}
      \item $f_{\mu}$ and $f_{{\sigma}^2}$: {\tt D128}
      \item $f_{\rm MLP}$: {\tt D512R-D512BR-D1024R}
     \end{itemize}
 \item $p(\mathbf{z}_1|\mathbf{z}_2)$ (Gaussian)
    \begin{itemize}
      \item $f_{\mu}$ and $f_{{\sigma}^2}$: {\tt D128}
      \item $f_{\rm MLP}$: {\tt D512R-D512R}
    \end{itemize}
 \item $p(\mathbf{x}|\mathbf{z})$ (Gaussian with fixed variance)
    \begin{itemize}
      \item $f_{\mu}$: {\tt DC3}
      \item $f_{\rm MLP}$: {\tt D4096R-DC256BR-DC128BR-DC64BR}
     \end{itemize}   
 \item $p(\mathbf{w}|\mathbf{z})$ (Gaussian with fixed variance)
    \begin{itemize}
      \item $f_{\mu}$: {\tt D10}
      \item $f_{\rm MLP}$: {\tt D512R-D4096R}
     \end{itemize}             
 \item $q(\mathbf{z}_2|\mathbf{z}_1)$ (Gaussian)
    \begin{itemize}
      \item $f_{\mu}$ and $f_{{\sigma}^2}$: {\tt D128}
      \item $f_{\rm MLP}$: {\tt (C64R-C128BR-C256BR-C256BR-F, D512R-D512BR)-D1024R-D64-D512R-D512R}
     \end{itemize}
\end{itemize}

Moreover, we set {\tt C64R-C128BR-C256BR-C256BR}
{\tt-F-D1024R-D1S} (where {\tt DkS} means {\tt Dk} with sigmoid function) as a network structure of the discriminator used in CelebA experiment.

\section{Conditional log-likelihood of JMVAE-h}
We can estimate conditional log-likelihood of JMVAE-h as follows.
\begin{eqnarray}
\log p(\tilde{\mathbf{x}}|\mathbf{w}) & \simeq & \log \int q(\mathbf{z}_2|\mathbf{w}) \int p(\mathbf{z}_1|\mathbf{z}_2)p(\tilde{\mathbf{x}}|\mathbf{z}_1) d\mathbf{z}_1 d\mathbf{z}_2 \nonumber \\
& \geq & \int q(\mathbf{z}_2|\mathbf{w}) \log \int p(\mathbf{z}_1|\mathbf{z}_2) p(\tilde{\mathbf{x}}|\mathbf{z}_1) d\mathbf{z}_1 d\mathbf{z}_2 \nonumber \\
& \geq & \int q(\mathbf{z}_2|\mathbf{w}) \int p(\mathbf{z}_1|\mathbf{z}_2) \log p(\tilde{\mathbf{x}}|\mathbf{z}_1) d\mathbf{z}_1 d\mathbf{z}_2 \nonumber \\
& \simeq & \frac{1}{N}\sum_{i=1}^N \int p(\mathbf{z}_1^{(l)}|\mathbf{z}_2) \log p(\tilde{\mathbf{x}}|\mathbf{z}_1) d\mathbf{z}_1,
\label{eq:hierarchical_loglike}
\end{eqnarray}
where $\mathbf{z}_2^{(l)}\sim q(\mathbf{z}_2|\mathbf{w})$ and 
\begin{eqnarray}
\int p(\mathbf{z}_1^{(l)}|\mathbf{z}_2) \log p(\tilde{\mathbf{x}}|\mathbf{z}_1) d\mathbf{z}_1 \simeq \frac{1}{N}\sum_{j=1}^N \log p(\tilde{\mathbf{x}}|\mathbf{z}_1^{(k)}),  \nonumber
\end{eqnarray}
where $\mathbf{z}_1^{(k)}\sim p(\mathbf{z}_1|\mathbf{z}_2^{(l)})$．We set $N=10$ for all experiments.

We find that the approximation of Eq. \ref{eq:hierarchical_loglike} holds a lower bound than that of JMVAE. Therefore, the evaluation value of the conditional likelihood of JMVAE-h might be underestimated compared to the evaluation value of JMVAE.

\end{document}